\documentclass{article}

\usepackage{arxiv}

\usepackage[utf8]{inputenc} 
\usepackage[T1]{fontenc}    
\usepackage{hyperref}       
\usepackage{url}            
\usepackage{booktabs}       
\usepackage{amsfonts}       
\usepackage{nicefrac}       
\usepackage{microtype}      
\usepackage{lipsum}
\usepackage{subfigure}
\usepackage{graphicx}
\usepackage{footnote}
\makesavenoteenv{tabular}
\makesavenoteenv{table}
\usepackage{caption}
\usepackage{subfigure}
\usepackage[ruled,vlined]{algorithm2e}
\usepackage{algorithmic}
\usepackage{float}
\usepackage{wrapfig}

\title{Training Multilingual Pre-trained Language Models with Byte-Level Subwords}

\author{
 Junqiu Wei, Qun Liu, Yinpeng Guo, Xin Jiang  \\
 Noah's Ark Lab, Huawei Technologies\\
   \texttt{\{weijunqiu, qun.liu, guo.yinpeng, jiang.xin\}@huawei.com} \\
}

\begin{document}
\maketitle

\begin{abstract}
The pre-trained language models have achieved great successes in various natural language understanding (NLU) tasks due to its capacity to capture the deep contextualized information in text by pre-training on large-scale corpora. One of the fundamental components in pre-trained language models is the vocabulary, especially for training multilingual models on many different languages. In the technical report, we present our practices on training multilingual pre-trained language models with BBPE: Byte-Level BPE (i.e., Byte Pair Encoding). BBPE has been adopted by pretrained language
models like GPT-2/3~\cite{radford2019language,brown2005language} and Roberta~\cite{Liu2019RoBERTaAR} and its usage in machine translation has been discussed in~\cite{wang2019neural}. 
We compared the byte-level vocabulary with the character-level vocabulary adopted in Google's multilingual BERT model through intensive case studies on the tokenization in a variety of languages. In the experiment, we adopted the architecture of NEZHA~\cite{wei2019nezha} as the underlying pre-trained language model and the results show that NEZHA trained with byte-level subwords consistently outperforms Google multilingual BERT and vanilla NEZHA by a notable margin in several multilingual NLU tasks. 
We release the source code of our byte-level vocabulary building tools and the multilingual pre-trained language models at the URLs \footnote{\url{https://github.com/huawei-noah/Pretrained-Language-Model/tree/master/BBPE}}\footnote{\url{https://github.com/huawei-noah/Pretrained-Language-Model/tree/master/NEZHA-TensorFlow}}.
\end{abstract}


\section{Introduction}
\label{sec:intro}

Pre-trained language models has demonstrated marvelous success by its excellent performance in a variety of natural languages understanding (NLU) tasks. In the pretraining phase, it employs language modeling tasks and learns contextualized word representations by utilizing the massive amount of training text. A large body of research efforts has been devoted to pre-trained language models such as ELMo~\cite{peters2018deep}, BERT~\cite{devlin2019BERT}, ERNIE-Baidu~\cite{sun2019ernie,sun2019ernie2}, ERNIE-Tsinghua~\cite{zhang2019ernie},
XLNet~\cite{yang2019xlnet}, RoBERTa~\cite{Liu2019RoBERTaAR}, NEZHA~\cite{wei2019nezha}, ALBERT~\cite{lan2019albert}, ELECTRA~\cite{clark2019electra} and MegatronLM\footnote{\url{https://nv-adlr.github.io/MegatronLM}}. As a fundamental technique in natural language processing (NLP), the language models pre-trained on text could be easily transferred to learn downstream NLP tasks with finetuning, which achieve the state-of-the-art performances on many tasks including sentiment analysis, machine reading comprehension, sentence matching, named entity recognition and natural language inference.
\begin{figure}
    \centering
    \includegraphics[width=0.8\textwidth]{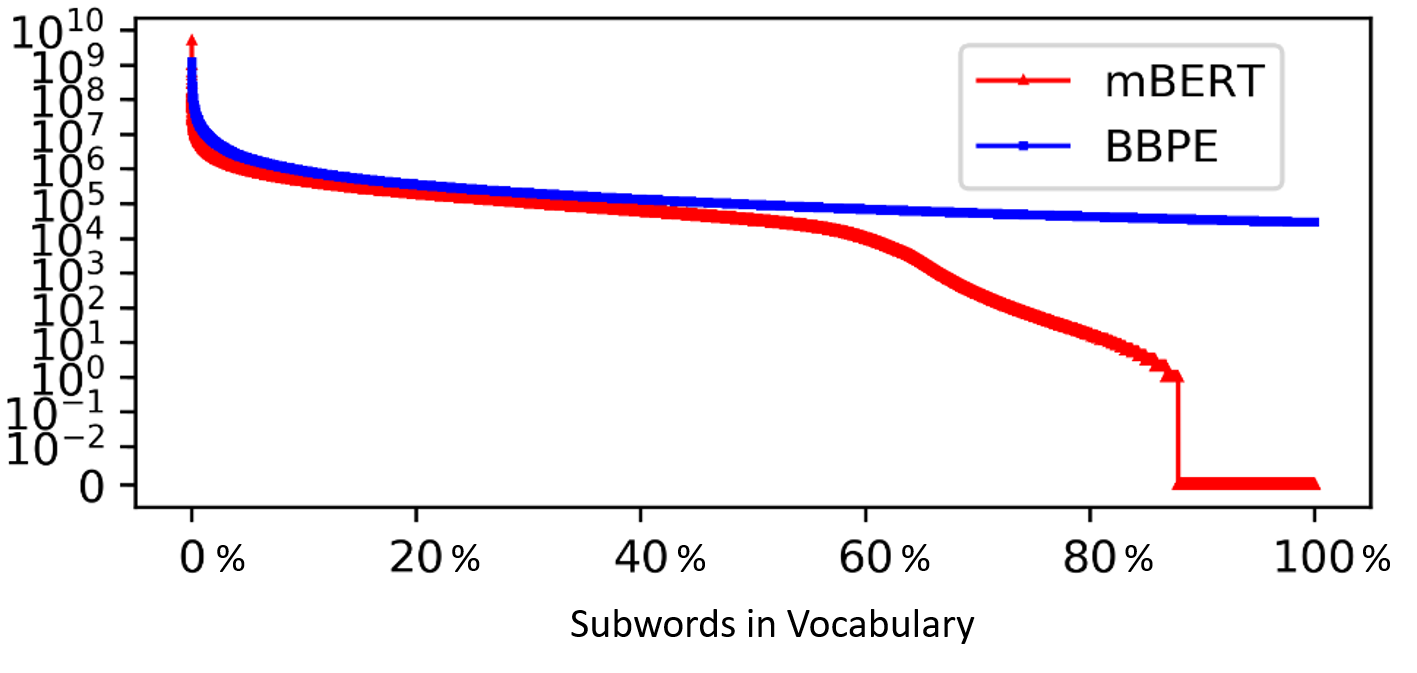}
    \caption{Subwords Frequencies in Google Multilingual Vocabulary (mBERT) and Our Byte-Level Vocabulary (BBPE)}
    \label{fig:vocab-freq}
\end{figure}

The multilingual pre-trained language model works as an universal cross-lingual encoder which embeds any sentence/words into a shared embedding space. The first attempt on this topic is Google's multilingual BERT~\cite{devlin2019BERT} which embeds more than 100 languages and highly improved the performance on low-resource languages. Then, \emph{XLM (Cross-Lingual Language Pre-training)}~\cite{conneau2019cross} further incorporates supervised parallel data (with a new cross-lingual language model objective) into the multilingual pre-trained language models. In this technical report, we detail our practice in training multilingual pre-trained language models with \emph{Byte-Level Subwords}. A key component in the pre-trained language models, esp. multilingual models, is the vocabulary. We observe that the character-level vocabulary, e.g., goolge multilingual BERT vocabulary (mBERT vocab), has two weaknesses. First of all, it has too many rare subwords contained whose representations are hard to learn in deep learning models. As shown in Figure~\ref{fig:vocab-freq}, the mBERT vocab has around 30\% subwords with the frequency less than 100 and roughly 18\% subwords which never occur. We calculate the frequencies of the subwords in the Wikipedia datasets with 14 languages as shown in Section~\ref{sec:moreexp} and sorted the subwords by its frequency. Note that we only consider the subwords which belong to the 14 languages in mBERT vocab. We observe that these rare subwords are mainly rare Chinese characters and rare Latin subwords. These subwords waste the slots of the vocabulary. Secondly, even though mBERT vocab has many rare subwords, it is impossible to avoid the unknown words problem (i.e., [UNK]). By the most up-to-date standard of Unicode~\footnote{\url{https://en.wikipedia.org/wiki/Unicode}}, there are more than 14 millions of characters (most of which rarely occur). And thus, it is not possible to include them all in a vocabulary which will lead to [UNK] problem in the text containing these non-included characters. For example "\raisebox{-0.1mm}{\includegraphics[scale=0.1]{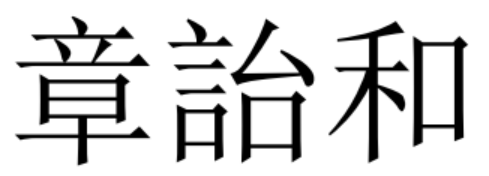}}" is the name of a famous Chinese writer
, however, the character "\raisebox{-0.35mm}{\includegraphics[scale=0.1]{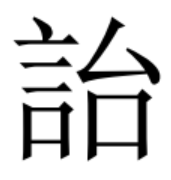}}" is not in the mBERT vocabulary. 

\begin{figure}
    \centering
    \includegraphics[width=\textwidth]{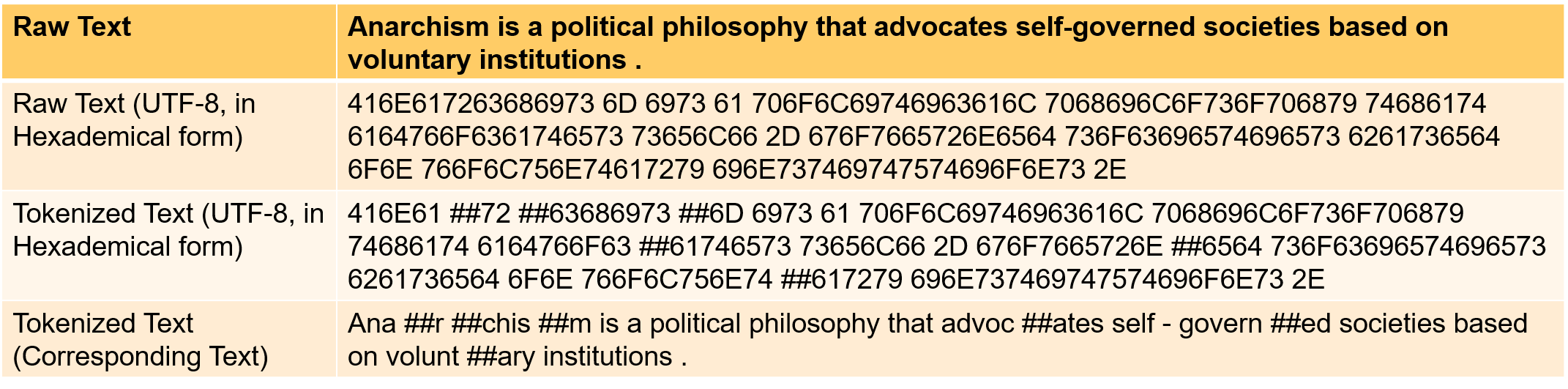}
    \caption{An Example of Tokenization with Byte-Level Subwords}
    \label{fig:example}
\end{figure}

Motivated by this, we employed a technique, namely \emph{Byte-Level Subwords} which shows marvelous success in neural machine translation~\cite{wang2019neural}, in building the vocabulary for multilingual pre-trained language models. Specifically, this technique first converts the text into its corresponding UTF-8 codes and then applies a byte-level vocabulary building algorithm on the UTF-8 codes. In this study, we employed \emph{Byte Pair Encoding (BPE)} as a show case of this underlying character-level vocabulary building algorithm and thus, it is called "Byte-Level BPE" (BBPE). There are strong experimental evidences~\cite{wang2019neural} showing that building the multilingual vocabulary in a byte-level will largely encourage the sharing of subwords among different languages in a fine-grain, as a result of which, the subwords obtained in the vocabulary have much higher frequencies and the learning of the representations of the subwords could be improved. As shown in Figure~\ref{fig:vocab-freq}, our BBPE vocab has all its subwords with the frequency more than 10,000 on the Wikipedia datasets of 14 languages. Besides, we thoroughly avoid the unknown words problem (i.e., [UNK]) by including all bytes into the vocabulary. It is worth mentioning that there are only 256 bytes in total and thus, the cost of including them all in a vocabulary is negligible. To intuitively give an idea of this Byte-Level Subwords, we demonstrate an example in Figure~\ref{fig:example}. In this figure, there are four lines which correspond to the raw text, the UTF-8 encoding of the text, the tokenized UTF-8 encoding of the text and the corresponding text of the tokenized UTF-8 encoding, respectively. Each word contains several characters and we convert each character into bytes and preserve the boundary of each word (i.e., whitespace). By applying a vocabulary building algorithm (e.g., BPE) to the UTF-8 encoding of the text, we obtain a byte-level vocabulary. The tokenization using the vocabulary is shown in the third line and the last line shows the corresponding text of the tokenized UTF-8 encoding. Note that the symbol "\#\#" denotes the subword is an trailing subword (i.e., the subword is not in beginning of the word it belongs to). In this work, we handcraft several techniques to better make sense of the byte-level subwords learned which will be presented in details in later sections. We also provide some insightful comments on these techniques which shed light on the practice of multilingual model training.

The contribution of this technical report is three-folds. First, we detail the deployment of the Byte-Level Subwords in the vocabulary training together with several effective handcrafted techniques integrated for the multilingual pre-trained language models. Secondly, we conducted comprehensive case studies on the tokenziation on multiple languages and demonstrate the effectiveness of the BBPE, esp. for the low-resource languages. Thirdly, we conducted experiment based on our previously released architecture NEZHA and our result shows that NEZHA with BBPE consistently outperform the vanilla NEZHA and Google multilingual BERT by a notable margin in natural language inference (XNLI) task. Fourthly, we open sourced our multilingual pre-trained language models and the BBPE tool. 

The remainder of this paper is organized as follows. In Section~\ref{sec:related}, we review the related works of our work. Section~\ref{sec:encoding} and Section~\ref{sec:exp} present our proposed byte-level vocabulary building algorithm and its empirical performance in training multilingual pre-trained language models. Finally, Section~\ref{sec:conl} concludes this paper.

\section{Related Works}
\label{sec:related}

This section reviews the literature related to our work.  Section~\ref{sec:rel:utf8}, Section~\ref{sec:rel:pretrained} and Section~\ref{sec:rel:vocab} present the existing works on the pre-trained language models and the vocabulary building algorithms, respectively. 

\subsection{Unicode and UTF-8 Encoding}
\label{sec:rel:utf8}

The unicode~\footnote{\url{https://en.wikipedia.org/wiki/Unicode}} is a unified standard for the consistent encoding, representation and handling of text data. Its success in unifying the representation of text renders it widely spread and applied in computer softwares. UTF-8 (8-bit Unicode Transformation  Format)~\footnote{\url{https://en.wikipedia.org/wiki/UTF-8}} is one of the most important formats in unicode. And UTF-8 is a variable-length character encoding and it encodes characters with 1-4 bytes. It is worth mentioning that UTF-8 was designed for backward compatibility with ASCII. Specifically, the first byte of UTF-8 are encoded using a single byte with the same binary value as ASCII. As such, each valid ASCII text is valid UTF-8-encoded Unicode as well. And the first byte of UTF-8 determines the length of the character encoding. Up to 2021, UTF-8 is by far the most commonly used encoding for the World Wide Web~\footnote{\url{https://w3techs.com/technologies/cross/character_encoding/ranking}}.

\subsection{Pre-trained Language Models}
\label{sec:rel:pretrained}

The pre-trained language models~\cite{peters2018deep,devlin2019BERT,sun2019ernie,sun2019ernie2,zhang2019ernie,yang2019xlnet,Liu2019RoBERTaAR,wei2019nezha,lan2019albert,clark2019electra} utilizes a massive amount of unsupervized text data in the pre-training phase and thus, the deep contextualized representations of words are obtained. In the finetuning phase, the representations could be successfully transferred to a wide range of NLP tasks including Text Classification, Natural Language Inference, Machine Comprehension, etc. The multilingual pre-trained language models~\cite{devlin2019BERT,conneau2019cross} works as a universal encoder which applies to many different languages and embedds them into a unified vector space. Compared with the monolingual models, the multilingual models enjoy the sharing of semantic information among different languages (esp., the language-agnostic part) and thus, highly boost the performance on low-resource languages. 

\subsection{Vocabulary Building Algorithms}
\label{sec:rel:vocab}

This section surveys the existing works of vocabulary building which has two categories, namely \emph{Character-Level Vocabulary} and \emph{Byte-Level Vocabulary}. The character-level vocabulary treats each character as the most fundamental unit in the vocabulary building and builds the vocabulary on the raw text representation of the corpus directly. But the byte-level vocabulary treats each byte instead as the most fundamental unit instead and build the vocabulary based on the byte representation of the corpus correspondingly. 

\subsubsection{Character-Level Vocabulary}
\label{sec:charVocab}

\textbf{Word Vocabularies. }This type of vocabularies were deployed in word-based Neural Machine Translation models in the earlier days. The word vocabularies consist of several words and simply treats each word as a token in the tokenization process~\cite{kalchbrenner2013recurrent,bahdanau2015neural,luong2015addressing,jean2015using}. But this method suffers from the unknown or rare words and highly discourages the subword sharing among different words and thus, become less popular in deep learning models which is vulnerable to unknown or rare words. 

\textbf{Character Vocabularies. }Due to the limitations of the word vocabularies, character vocabularies are deployed which contain several characters and treat each character as a token in the procedure of text segmentation~\cite{al2019character}. The advantages of this type of vocabularies are two folds. First, it has very small size and thus, could reserve much more GPU memory for storing more training samples (i.e., the batch size could be increased). Secondly, it highly encourages the sharing of subword information among different words. Despite its advantages, its limitation is also obvious. The character vocabularies lead to much longer sequences of tokens after the tokenization which renders the training of the models harder. 

\textbf{Subword Vocabularies. }The subword vocabularies consist of several subwords and thus, properly balanced the word vocabularies and the character vocabularies and becomes de facto standard of building vocabularies. There are three major subword vocabulary building algorithms, namely \emph{Byte Pair Encoding (BPE)}~\cite{bpe,sennrich2016neural}, \emph{Unigram}~\cite{kudo2018subword} and \emph{WordPiece} in the subword vocabulary building. BPE and WordPiece initialize the vocabulary as a set of characters and then iteratively merges a pair of tokens in the vocabulary and insert the merged pair (i.e., a newly created token) into the vocabulary until the vocabulary size reaches a predefined value. Their difference lies on their selection method of the token pair in each iteration. 
BPE iteratively replaces the most frequently occurred pair of consecutive characters in a text with a single character that does not occur in the text. Besides, it maintains a mapping table to link each pair replaced to their corresponding character for the usage of decoding. In the vocabulary training, each character is treated as the most fundamental unit. 
Consider the example as shown in Table~\ref{tab:bpe}. The raw text is "ABABABCABC" and the most frequent pair of characters is "AB". Thus, in the first iteration, the pair "AB" is replaced with a new character Z and "Z=AB" is inserted into the mapping table (i.e., vocabulary). Similarly, in the last three iterations, "ZC", "ZZ", "YY" are replaced by "Y", "X" and "M", respectively.  It is worth mentioning that the whitespace is not allowed to be merged with any other character. This means that the characters merged in any iteration must be within the same word and each item in the vocabulary must be a sub-string of a word. WordPiece is the same as BPE except the selection of the character pair in each iteration. WordPiece selects the one which maximizes the likelihood of a given language model after the merge of the pair. In a word,  BPE and WordPiece construct the vocabulary in a bottom-up fashion which start from character-level vocabulary and iteratively enrich the vocabulary by merging two existing tokens.

\begin{table}[ht]
\caption{An Example of BPE}
\label{tab:bpe}
\begin{center}
\begin{tabular}{|c|c|c|}
\hline
 & Text & Mapping Table (Vocabulary)\\ \hline
Raw Text         &  ABABABCABC    &  - \\ \hline
 Iteration 1  & ZZZCZC & Z=AB  \\ \hline
Iteration 2 & ZZYY & Z=AB, Y=ZC \\ \hline
Iteration 3 & XYY & Z=AB, Y=ZC, X=ZZ \\ \hline
Iteration 4 & XM & Z=AB, Y=ZC, X=ZZ, M=YY \\ \hline
\end{tabular}
\end{center}
\end{table}

Conversely, Unigram constructs the vocabulary in a top-down fashion which starts from a set of words/subwords and enriches the vocabulary by splitting the existing tokens instead of merging them. Specifically, it initially maintains a set of candidates of tokens (typically, all words in corpus) and then iteratively splits the candidates by a probabilistic model and insert the split ones into the candidate list until the candidate list reaches a certain size. Several pruning techniques are also incorporated to prune the candidates and boost the building of the vocabulary. BPE and Unigram are integrated into the renowned tool, namely \emph{SentencePiece} library~\cite{kudo-richardson-2018-sentencepiece}. But the implementation of WordPiece which is developed by Google has not been released due to commercial issues and is not available. The subword vocabulary was first deployed to the field of \emph{Neural Machine Translation} for training vocabulary~\cite{sennrich2016neural} and was later on introduced to pre-trained language models. BERT~\cite{devlin2019BERT} uses WordPiece as their underlining vocabulary building algorithm and ERNIE-Baidu, ERNIE-Tsinghua, NEZHA, ELECTRA simply use the vocabulary of BERT released by Google in training their models. ALBERT~\cite{lan2019albert} and XLNET~\cite{yang2019xlnet} use Unigram and BPE respectively by using the SentencePiece library~\cite{kudo-richardson-2018-sentencepiece}. 

\subsubsection{Byte-Level Vocabularies}
\label{sec:byteVocab}

As shown in Figure~\ref{fig:example}, the second line shows the byte-level text (in UTF-8 codes) of the original sentence shown in the first line. Byte-Level vocabularies are built and applied in the converted byte-level text~\cite{radford2019language,Liu2019RoBERTaAR,wang2019neural}. In the byte-level representation, each character in the original text is converted to multiple bytes in the byte-level text and thus, byte-level vocabularies allow the sharing of different words in a finer grain (i.e., in a sub-character level), as a result of which, the vocabularies obtained have less rare tokens and more high-frequency tokens (e.g., each rare Chinese/Japanese/Korean character is tokenized into multiple high-frequency bytes). ~\cite{wang2019neural} deploys the byte-level BPE on the neural machine translation systems and GPT-2
/3~\cite{radford2019language,brown2005language} and RoBERTa~\cite{Liu2019RoBERTaAR} claims their vocabularies are constructed by using BPE over byte-level text. But they did not provide the details of their vocabulary building procedure. A vocabulary consisting purely of 256 bytes has also been used in many language and speech tasks such as part-of-speech tagging and named entity recognition~\cite{gillick2016multilingual}, translation~\cite{ruiz2017byte}, machine reading~\cite{kenter2018byte} and speech recognition~\cite{li2019bytes} but these work did not adopt BPE in text encoding.

\section{Building Vocabulary with Byte-Level Subwords}
\label{sec:encoding}

In this section, we present our practice on building the vocabulary with byte-level subwords in details. Section~\ref{sec:vocabbuilding} presents the overall byte-level vocabulary algorithm step by step. Section~\ref{sec:tricks} provides the insights on the technical details involved in our algorithm to give a high level idea. Section~\ref{sec:discussion} discusses from the theoretical perspective the reasons why byte-level subwords work and when this technique works.


\subsection{Vocabulary Building Algorithm}
\label{sec:vocabbuilding}

In this section, we present the details of our byte-level vocabulary building algorithm and provide the insights on the important tricks deployed which sheds light on the practice of building the byte-level subwords. 
It is worthy mentioning that in this section, we use BPE as our underlining vocabulary building tool on the byte-level text but our algorithm is compatible with any other vocabulary building tool (character-level algorithm mentioned in Section~\ref{sec:charVocab}) such as Unigram, WordPiece etc. (i.e., the BPE mentioned in this section could be replaced with any other character-level vocabulary building algorithm in Section~\ref{sec:charVocab} such as WordPiece, etc.). 
Since we consider the text as a byte sequence, we call the training method \emph{Byte-Level Byte Pair Encoding (BBPE)}. Note that we assume that our algorithm takes the raw text as input and thus, the vocabulary building mentioned in this section contains the procedure of converting the raw text into byte sequence and applying BPE to the byte-level text. 
In the remainder of this paper, we refer the term "word" to a sequence of characters/bytes which is allowed to be merged in BPE (i.e., the sequence has no whitespace within but has a whitespace before and after itself). 

\begin{algorithm}[H]
\SetAlgoLined
\KwData{Raw Multilingual Text Data}
\KwResult{A Vocabulary Containing Byte-Level Subwords}

\textbf{Step 1: Text Preprocessing}\;
\begin{itemize}
\item Step 1.1: Insert Whitespace before and after each CJK character and each punctuation\;

\item Step 1.2: Convert each character into its corresponding UTF-8 byte sequence except for whitespaces and sentence stops. This step converts the raw text into byte-level representation. But we keep whitespaces and sentence stops intact which provides the boundary information of each word and sentence, respectively\;

\item Step 1.3: Assign first byte in each converted byte-level word a "leading" label denoting it is the first byte of a word\;
\end{itemize}

\textbf{Step 2: Apply BPE to the byte-level text}\;

\begin{itemize}
    \item This step applies BPE as mentioned in Section~\ref{sec:charVocab} to the byte-level text given by Step 3 and obtain a vocabulary containing byte-level subwords. 
\end{itemize}

\textbf{Step 3: Vocabulary Postprocessing}\;

\begin{itemize}
\item Step 3.1: Insert all bytes (totally 256) into vocabulary if they are not in the vocabulary\;

\item Step 3.2: Insert all bytes (totally 256) each of which is assigned a "leading" label into vocabulary if they are not in the vocabulary\;


\item Step 3.3: Insert "\#\#" before each trailing subword and remove the labels added in Step 1.3 and Step 3.2. 

\end{itemize}

\label{algorithm:vocabbuilding}
 \caption{Vocabulary Building Algorithm with Byte-Level BPE}
\end{algorithm}

We proceed to present the step-by-step description of the algorithm which consists of the following three steps as shown in Algorithm~\ref{algorithm:vocabbuilding}. 
Step 1 preprocesses the raw input data and converts it into its corresponding byte-level representation. It involvs three sub-steps which (1) insert whitespaces before and after each CJK (Chinese, Japanese or Korean) character and each punctuation, (2) covert each character into its UTF-8 byte codes (except for whitespaces and sentence stops) (3) assign a label to each leading byte (as shown in Step 1.1, Step 1.2 and Step 1.3). Step 2 applies BPE to the byte-level text and finally obtains a vocabulary containing byte-level subwords. Step 3 postprocesses the vocabulary to make sure (1) the vocabulary has 256 single-byte leading subwords and 256 single-byte trailing subwords (2) each trailing subword has the symbol '\#\#' in its beginning.

\subsection{Technical Details for Building A Byte-Level Vocabulary}
\label{sec:tricks}

In this section, we present three technical details deployed in our BBPE algorithm and provide the insights on each one as follows. 

\subsubsection{Basic Units: Bytes}

\textbf{Treat each byte as the basic unit. }The UTF-8 encoding represents each unicode character with 1-4 bytes. Following~\cite{wang2019neural}, in our algorithm, we consider the UTF-8 encoding of the text and the basic unit used is byte. This decomposes each character into multiple pieces (i.e., bytes) and thus, allows all words to share the subwords in a finer level (i.e., up to sub-character level) and also have more chances of sharing. Besides, it decomposes the rare words/characters (esp. characters) into several bytes with a fairly high frequencies so that pre-trained language models could learn their contextualized representations well enough. We will see in Section~\ref{sec:tokenization} through case studies that even we split each character into bytes, the unicode of most character could be merged as a subword in our BBPE vocabulary. And the unicode of each rare character is split into multiple subwords in BBPE. On the other hand, we do not break each byte into more detailed pieces (i.e., bits) since byte-level is fine enough (note that there are only 256 distinct bytes). 

\textbf{Vocabulary must contain all single-byte subwords (in both leading and trailing versions mentioned in Section~\ref{sec:distinguish}). }Following~\cite{wang2019neural}, in case that the corpus in the pre-training or finetuning phase contains unseen words/characters that never occur in the corpus for training vocabulary, we make sure that the vocabulary contains all single-byte characters (in both versions aforementioned and thus, there are totally 512 single-byte subwords in our vocabulary). As a result, we could always decompose any given sequence of a UTF-8 characters into a sequence of tokens contained in our vocabulary (i.e., there will be no unknown tokens in the tokenization results with our vocabulary). This trick corresponds to Step 3.1 and Step 3.2 which ensure the vocabulary contains all single-byte subwords in both version and as such, the unknown token problem will be avoided. 

\subsubsection{Maximum Units: Words, CJK Characters and Punctuations}

\textbf{The UTF-8 encoding of each word is treated as a maximum unit in BBPE}. Consider a word in the original text which has a whitespace before and after itself. We keep the boundary of the word intact (i.e., whitespaces) when converting the raw text into UTF-8 codes. The same as the vanilla BPE, we do not allow the cross-word merges and assume that word is the largest token to be used, as a result of which, it reduces the occurrences of long tokens which normally has low frequencies. The languages such as Latin (including English, French, Italic) and Slavic (including Russian, Polish) naturally has the boundary for each word and we preserve the word boundary (i.e., denoted by whitespace) in the data preprocessing. This trick corresponds to Step 1.2 which preserves the word boundary in the raw text. This trick was also deployed in~\cite{wang2019neural} in the byte-level neural machine translation. 

\textbf{The UTF-8 encoding of each CJK (Chinese, Japanese, Korean) character and each punctuation is also treated as a maximum unit. }Different from Latin languages, the raw text of CJK does not have the information of the boundaries of words and each sentence in CJK is simply a sequence of characters (without whitespace contained). Consider that there are ambiguities on the segmentation (i.e., tokenization) of CJK text and a poor segmentation on the text could lead to significant performance loss on many downstream tasks, esp. Named Entity Recognition (NER), in pre-trained language models such as BERT~\cite{devlin2019BERT}, RoBERTa~\cite{Liu2019RoBERTaAR}, etc., it is a common practice that each CJK character is treated as a token. We also follow this practice and simply treat each CJK character as a word in the vocabulary training and safely let the pre-training language models learn the word-level/phrase-level information (i.e., the dependencies among characters) with the attention mechanism. Besides, in the non-CJK text, each number such as ``233000'', ``2019'' is a word (i.e., surrounded by two whitespaces before and after it) but in the CJK text, each number is not separated with other characters. Thus, another benefit of this trick is to make each number in CJK text separated from other text and treated as a word. The trick corresponds to Step 1.1 which ensures that the bytes inside each CJK character (punctuation) will not be merged with any byte outside this CJK character (punctuation).

\subsubsection{Distinguish Trailing Subwords from Leading Subwords}
\label{sec:distinguish}

Consider a subword sequence \{"whats", "app"\}. Its original text has two possibilities: "whats app" and "whatsapp" if we do not distinguish a trailing subword and a leading subword with the same spelling. Thus, we observe that although "app" and "\#\#app" have the same spelling, their semantic information is different. Besides, another benefit of this practice is to make tokenization lossless. This means that after a sentence is tokenized into several tokens, we could recover the original sentence by using the symbol ("\#\#") denoting an trailing subword. Motivated by this, we distinguish the two cases in our algorithm. It is worthy mentioning that this is a commonly used technique in Neural Machine Translation~\cite{sennrich2016neural,wang2019neural} which we also deploy in this report. In Section~\ref{sec:expwithout}, we also conduct ablation study on the BBPE without using this technique. This technical detail corresponds to Step 1.3 which ensures that we could distinguish a trailing subword and a leading subword with the same spelling. We will show in the experiment that this technique is highly effective in training multilingual pre-trained language models which implies that a trailing subword and a leading subword have different semantic meanings. 

\subsection{Discussion on Why and When Byte-Level Subwords Work}
\label{sec:discussion}

It is a common sense that deep neural networks for natural language processing, esp. the pre-trained language models which has millions or even trillions of parameters, are vulnerable to rare and unknown words. This is because the scarcity of the rare words renders the learning of their representations quite hard in deep neural networks since they are rarely exposed to the model in the training procedure and waste quite a lot of slots in the vocabulary. And the unknown words are simply treated as a special token such as '[UNK]' without further distinguishing the spelling. This problem is especially severe in character-rich languages such as Thai, Arabic, Japanese, etc. Representing the text in byte-level is an effective solution since each original character is converted into 1-4 bytes and thus, the sub-character level sharing among different words or original characters are possible. With the byte-level subwords, one original rare or unknown character could be split into several frequent bytes and equivalently speaking, the slots of the rare words in the vocabulary could be freed for storing more frequently used and more meaningful symbols. Thus, the problem of rare/unknown tokens is largely mitigated. And the byte representation is language agnostic and even allows the sharing between languages without any overlap on their character sets. 

By the above discussion, we could obtain that the byte-level subwords works in the scenarios where (1) there are rare characters in the character-level text (esp. character-rich languages such as Thai, Arabic, Japanese, etc.) (2) the rare character has more than 1 bytes in the byte representation (so that it could be decomposed into more than one bytes which occur frequently in the training corpus). As we will present in the experiment, byte-level subwords get marvelous success in the languages whose character corresponds to multiple bytes and have a large number of characters but obtain almost ignorable improvement on Latin languages (since Latin languages has very limited number of characters and the UTF-8 unicode of most character in Latin have only 1 byte) compared with character-level subwords. This result confirms our discussion in this section. 

Finally, it is worth mentioning that in the BBPE vocabulary, it is possible that a byte-level token has no corresponding original text (e.g., a sub-string of a UTF-8 unicode of a Chinese character which contains more than 1 bytes). This will not lead to any problem in the natural language understanding as studied in this paper. But in the natural language generation tasks, it may result in infeasible word in the decoding procedure. But this problem could be highly alleviated by many effective solutions such as data augmentation which reduces the probability of generating infeasible words. 

\begin{figure}
    \centering
    \includegraphics[width=\textwidth]{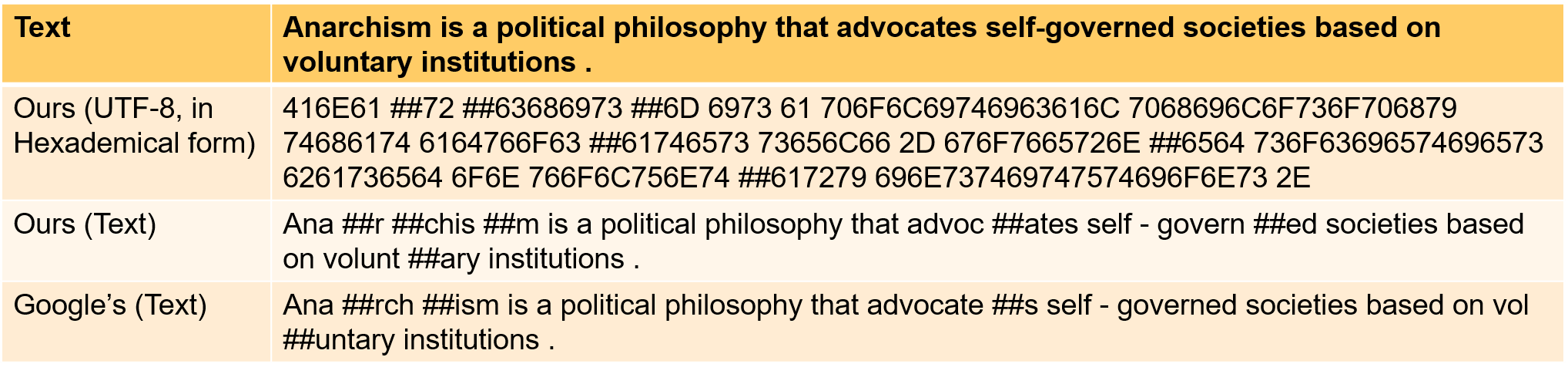}
    \caption{Case Study on English}
    \label{fig:english}
\end{figure}

\begin{figure}
    \centering
    \includegraphics[width=\textwidth]{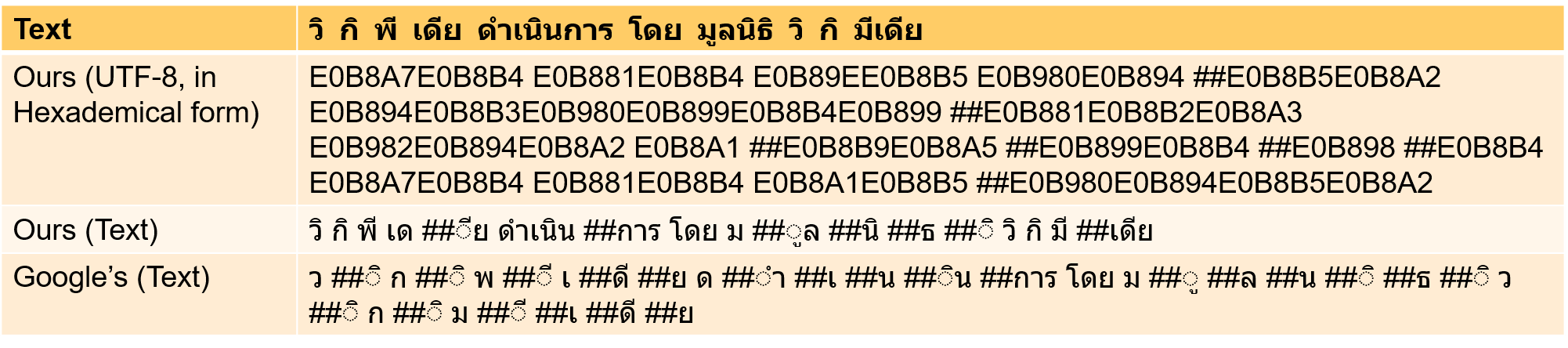}
    \caption{Case Study on Thai}
    \label{fig:thai}
\end{figure}

\begin{figure}
    \centering
    \includegraphics[width=\textwidth]{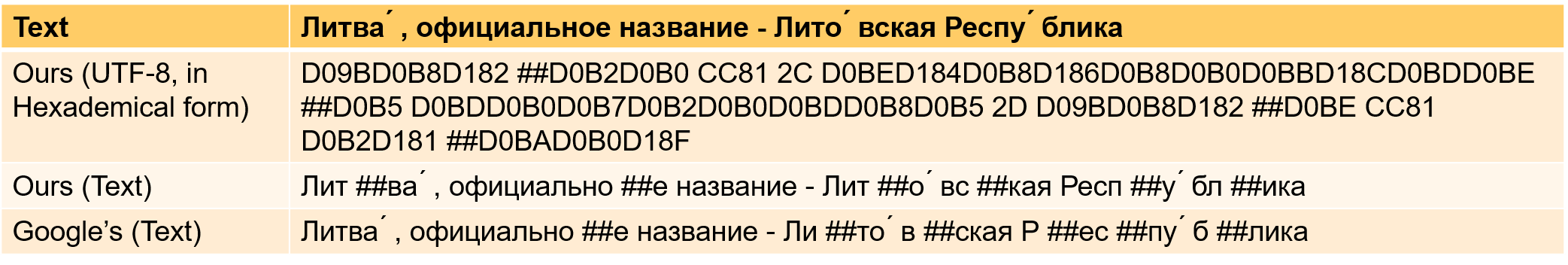}
    \caption{Case Study on Russian}
    \label{fig:russian}
\end{figure}

\begin{figure}
    \centering
    \includegraphics[width=\textwidth]{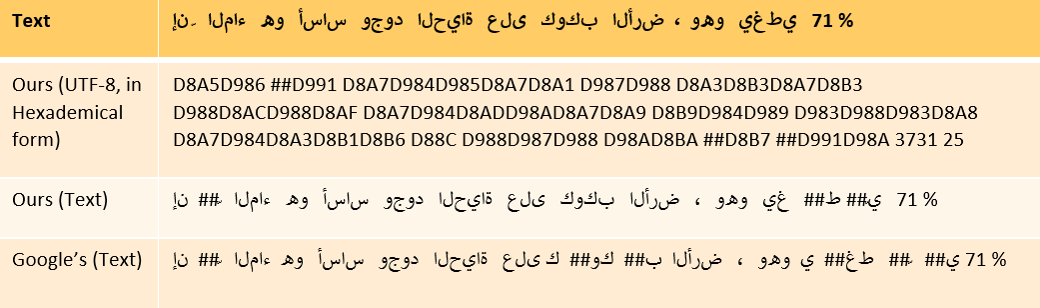}
    \caption{Case Study on Arabic}
    \label{fig:arabic}
\end{figure}

\begin{figure}
    \centering
    \includegraphics[width=\textwidth]{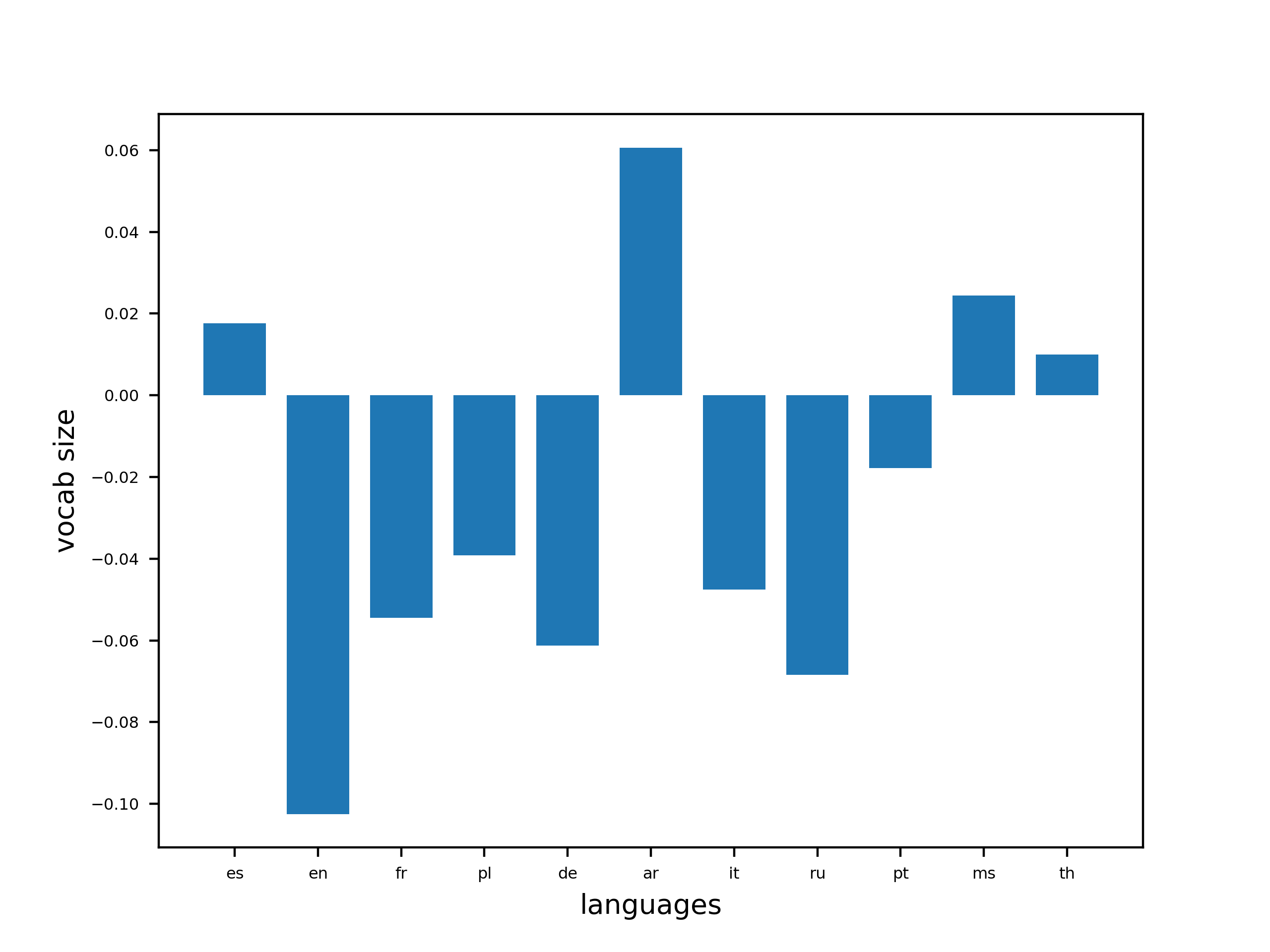}
    \caption{Comparison between Google Vocabulary and BBPE Vocabulary}
    \label{fig:11lang}
\end{figure}

\section{Experiments}\label{sec:exp}

In this section, we report the experimental results of our methods. The remainder of this section is organized as follows. Section~\ref{sec:expsetting} provides the information of our experimental settings. Section~\ref{sec:tokenization} presents the case studies (i.e., the tokenization) on many different languages by using our vocabulary and comparison with the vocabularies used in other works. Section~\ref{sec:exp_result} presents the experimental results on the multilingual XNLI tasks. Finally, Section\ref{sec:abalation} further studies the effect of several variants of our vocabulary building algorithm and also presents the empirical study on the experiments with more languages. 

\subsection{Experimental Setting}
\label{sec:expsetting}

\paragraph{Pre-training Details} We train the NEZHA models on 10 servers on Huawei Cloud~\footnote{\url{https://www.huaweicloud.com/product/modelarts.html}}, each of which has 8 NVIDIA Tesla V100 GPUs with 32GB memory. The distributed training algorithm is the \emph{Ring-AllReduce}\footnote{\url{https://github.com/baidu-research/baidu-allreduce}} and was employed with the framework named Horovod~\cite{sergeev2018horovod}. We trained each model from scratch and pre-train for 80k steps. 
The batch size on each GPU is 64, and thus the total batch size is 64 * 8 * 10 = 5120. 
For the each model tested, we set the maximum learning rate to be $1e-4$ (with 1800 warm-up steps and polynomial decay). 
In addition, we adopted the mixed-precision training using FP16~\cite{micikevicius2017mixed} in the pre-training phase. 
We employ wikipedia datasets containing 11 languages including English, Arabic, Spanish, Thai, Russian, German, Italian, French, Malay, Portuguese and Polish and upsample the low-resource languages to make the dataset more balanced. 

\begin{figure}
    \centering
    \includegraphics[width=\textwidth]{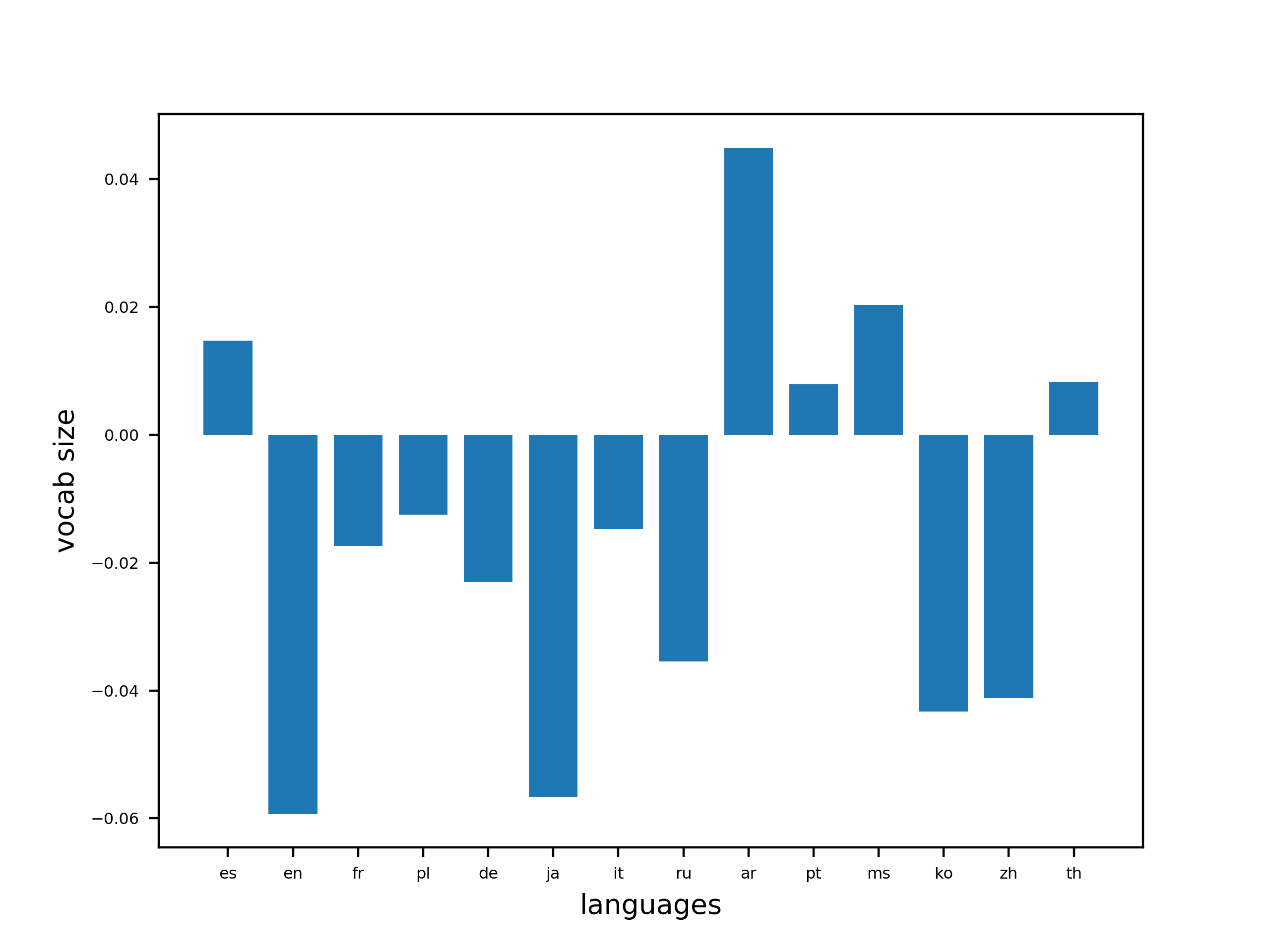}
    \caption{Comparison between Google Vocabulary and BBPE Vocabulary (on 14 Languages)}
    \label{fig:14lang}
\end{figure}

\subsection{Case Studies On Tokenization Results and Vocabulary Analysis}
\label{sec:tokenization}

Figure~\ref{fig:english}-Figure~\ref{fig:arabic} demonstrates the tokenization of many different languages by using our vocabulary and that used in Google multilingual BERT. Our BBPE vocabulary is built by using wikipedia datasets containing 11 languages including English, Arabic, Spanish, Thai, Russian, German, Italian, French, Malay, Portuguese and Polish and upsample the low-resource languages to make the dataset more balanced. Each figure has four lines which corresponds to raw text, the UTF-8 encoding of the raw text converted by the aforementioned 6 steps, the tokenization result on the UTF-8 encoding, the corresponding text of the tokenized UTF-8 encoding and the tokenization result on the raw text by using Google's vocabulary. From Figure~\ref{fig:english},  Figure~\ref{fig:thai} and Figure~\ref{fig:arabic}, we observe that for the same sentence in English, Thai or Arabic, the output of our tokenizer has less tokens and this implies that our deployed BBPE enjoys better sharing of subwords. 

Figure~\ref{fig:11lang} shows the language-wise comparison between Google multilingual vocabulary and our BBPE vocabulary. Each bar in the figure shows the relative difference of the tokens in the corresponding language (i.e., (A - B) / A, where A (B resp.) is the number of tokens in the language in BBPE (Google multilingual resp.) vocabulary). As observed from the figure, BBPE vocabulary has more tokens in Spanish, Arabic, Malay and Thai and this is the reason why our BBPE vocabulary gives less tokens in the tokenization of Arabic and Thai compared with Google multilingual vocabulary. And we also observe that although BBPE vocabulary has less tokens in English and Russian, it has almost the same tokenization result on the two languages as that of Google vocabulary. This implies that our BBPE vocabulary removes redundant tokens (esp., rare words/subwords) in languages like English, Russian, French etc. and reserves the tokens for low-resource languages like Arabic and Thai. 
Consider again the Chinese word "\raisebox{-0.1mm}{\includegraphics[scale=0.1]{zhangyihe.png}}" mentioned in Section~\ref{sec:intro} which is the name of a famous Chinese writer
, however, the character "\raisebox{-0.35mm}{\includegraphics[scale=0.1]{yi.png}}" is not in the mBERT vocabulary and will be treated as "[UNK]" (i.e., unknown token). But with our BBPE vocabulary, the UTF-8 encoding of the character (i.e., E8A992) will be tokenized into "E8", "\#\#A9" and "\#\#92" and the unknown word problem is properly avoided. 
Consider another example "\raisebox{-0.35mm}{\includegraphics[scale=0.7]{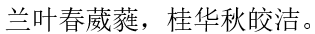}}" which is one sentence from an ancient Chinese poem. The tokenization result of mBERT is "\raisebox{-0.35mm}{\includegraphics[scale=0.7]{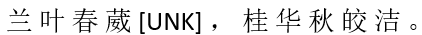}}", in which the word "\raisebox{-0.35mm}{\includegraphics[scale=0.7]{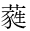}}" is not in its vocabulary and treated as "[UNK]". But the tokenization result of our BBPE is "E585B0 E58FB6 E698A5 E891 \#\#B3 E895 \#\#A4 EFBC8C E6A182 E58D8E E7A78B E79A \#\#8E E6B481 E38082" where the word "\raisebox{-0.35mm}{\includegraphics[scale=0.7]{cn-unkword.PNG}}" is tokenized into two subwords "E895" and "\#\#A4" and the "[UNK]" problem is properly circumvented again. Besides, "\raisebox{-0.35mm}{\includegraphics[scale=0.7]{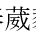}}" and "\raisebox{-0.35mm}{\includegraphics[scale=0.7]{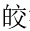}}" are two rare Chinese characters but they are included in the Google mBERT vocabulary which occupy two entries of the vocabulary and wastes the space. Our BBPE vocabulary does not contain the two rare characters which reserves the space of vocabulary for more frequent subwords and each of the two characters is split into two subwords with higher frequencies in the tokenization. 

\subsection{Experimental Results}
\label{sec:exp_result}

We tested our methods and several baselines on \emph{Cross-Lingual Natual Language Inference (XNLI)} task on seven languages, namely English (En), Arabic (Ar), German (De), Spanish (Es), French (Fr), Russian (Ru) and Thai (Th). We trained two BBPE-based vocabularies with different sizes 50k and 100k, respectively and trained NEZHA with the two byte-level vocabularies. And we compared with Google's multilingual BERT, vanilla NEZHA (trained with the same vocabulary as Google's multilingual BERT) and NEZHA trained with original BPE-based vocabulary. The result is as shown in Table~\ref{tab:ta}. As the table shows, NEZHA (BBPE) with 100k-size vocabulary consistently outperforms BERT (Google), vinilla NEZHA and NEZHA (BPE) on the seven languages tested by a notable margin. The improvements of our NEZHA (BBPE) over BERT (Google) and NEZHA (BPE) are more significant on Ar, Ru and Th which are low-resource languages and this experiment results demonstrate that these languages enjoys the byte-level subword sharing. And this results are also consistent with the case studies of tokenization as shown in Figure~\ref{sec:tokenization} (our BBPE gives less tokens in the tokenization results on these languages compared with those by using Google multilingual vocabulary). The improvement brought by BBPE on the Latin languages such as English is not very significant since most Latin character is single-byte character and there is no difference between character-level and byte-level. 
Besides, NEZHA (BBPE) which contains only 50k subwords in its vocabulary (i.e., less than or equal to half of the Google's vocabulary and that of NEZHA (BPE)) outperforms BERT (Google) and NEZHA (BPE) and its performance is very close to vinilla NEZHA.

\begin{table}[ht]
\caption{Results on \emph{XNLI}}
\label{tab:ta}
\begin{center}
\begin{tabular}{|c|c|c|c|c|c|c|c|c|c|}
\hline
Model &  En     & Ar     & De     & Es     & Fr     & Ru     & Th   & Avg.  & Vocab\\ \hline
BERT  (google)          &       82.4    &72.0    &76.1    &78.4    &76.9    &74.5    &67.1  & 75.3  & ~120k  \\ \hline
 NEZHA  &83.0    &75.9    &79.1    &81.1    &79.3    &76.3    &68.8  & 77.6  & ~120k  \\ \hline
 NEZHA (BPE) &81.2  & 72.7 &77.0  &78.8  &79.0  &73.8  &67.3 & 77.0 & ~100k \\ \hline 
NEZHA (BBPE)& 83.2 & 73.9&79.8 &81.5 &80.8 &77.1 &71.5 & 78.3 & ~100k  \\ \hline
NEZHA (BBPE) &82.6 &74.1 &78.3 & 80.5 &79.6 &76.1 &69.6 & 77.3 & ~50k   \\ \hline
\end{tabular}
\end{center}
\end{table}

\subsection{Abalation Study}
\label{sec:abalation}

\subsubsection{Comparison between BBPE and BUnigram}
\label{sec:ablation}

In this section, we empirically compare two different word segmentation algorithms on the segmentation/tokenization of byte-level text. One is BPE which we adopted in our experiment and the other one is \emph{Unigram}~\cite{kudo2018subword}. The two algorithms are commonly used for word segmentation and are integrated in a renowned tool, namely \emph{SentencePiece}~\footnote{\url{https://github.com/google/sentencepiece}}. Since they both handle byte-level text, we call them \emph{BBPE} and \emph{BUnigram}, respectively. It is worth mentioning that BUnigram is space-consuming and could not affort to process all multilingual corpora. Thus, for each of the 11 languages we adopted for the pre-training, we sampled a subset of it and constructed a small version of the pre-training corpora. The result is as shown in Table~\ref{tab:bunigram}. From the result, we could observe that BBPE consistently ourperforms BUnigram on the seven languages tested. 

\begin{table}[ht]
\caption{Comparison between BBPE and BUnigram (Based on a Small Version of Data)}
\label{tab:bunigram}
\begin{center}
\begin{tabular}{|c|c|c|c|c|c|c|c|c|}
\hline
Model &  En     & Ar     & De     & Es     & Fr     & Ru     & Th     & Vocab\\ \hline

 NEZHA (wiki sample, BBPE) & 80.3 & 67.6 & 75.3 & 78.1 & 76.6 & 72.6 & 69.5 & ~100k
\\\hline
NEZHA (wiki sample, BUnigram) & 78.5 & 60.6 & 72.4 & 76.6 & 76.2 & 62.3 & 53.7 & ~100k \\\hline
\end{tabular}
\end{center}
\end{table}

\subsubsection{Comparison Between BBPE and That without Distinguishing A Leading and A Trialing Subword with Same Spelling}
\label{sec:expwithout}

In this section, we compare the experimental results on BBPE without using the technique in Section~\ref{sec:distinguish} (we refer this version as BBPE, w/o D in short) and our original BBPE presented before. We consider three different implementations of BBPE, w/o D which are shown as follows. 

\begin{itemize}
    \item (BBPE, w/o D, WWB (Without Word Boundary)): in the building of vocabulary, we do not employ the technique in Section~\ref{sec:distinguish}. 
    \item (BBPE, w/o D, WBL (Word Boundary Label)): in the building of vocabulary, we do not employ the technique in Section~\ref{sec:distinguish} but in the tokenization, we add a special character "\#\#" before each trailing subword. As a result, each trailing subword will have a different context compared with the leading subword with the same spelling. 
    \item (BBPE, w/o D, STE (Subword Type Embedding)): in the building of vocabulary, we do not employ the technique in Section~\ref{sec:distinguish} but assign a trainable subword type embedding for each type of subword which is added with word embedding in the input of the model. In this work, we consider 7 different subword types: trailing subword, leading subword, ending subword, whole word, "[SEP]", "[CLS]" and others. 
\end{itemize}

Table~\ref{tab:woP4} demonstrates the experimental results. The above three methods have worse performances compared with our original BBPE method. This implies that the explicit information of leading/trailing subwords in the vocabulary contains substantial semantic information and has impact on the performance. It is notable that the BBPE, w/o D, WBL leads to significant performance drop. This is because the additional character "\#\#" introduced make the sentence longer and the introduced characters are not informative for distinguishing leading and trailing subwords as expected. 

\begin{table}[ht]
\caption{Comparison with BBPE without using the technique in Section~\ref{sec:distinguish}}
\label{tab:woP4}
\begin{center}
\begin{tabular}{|c|c|c|c|c|c|c|c|c|c|}
\hline
Model &  En     & Ar     & De     & Es     & Fr     & Ru     & Th   & Avg.  & Vocab\\ \hline
NEZHA (BBPE)& 83.21 & 73.95&79.80 &81.50 &80.78 &77.09 &71.54 & 78.3 & ~100k  \\ \hline
NEZHA (BBPE, w/o D, WWB) & 82.48 & 72.51 & 78.96 & 80.10 & 80.32 & 77.27 & 72.63 & 77.75 & ~100k \\ \hline
NEZHA (BBPE, w/o D, WBL)& 79.56 & 65.07 & 77.69 &78.04&76.39&72.79&69.90 & 74.21 & ~100k \\ \hline
NEZHA (BBPE, w/o D, STE)& 82.93 & 72.95& 79.13 &80.54&80.44 &76.83 &73.95& 77.94 & ~100k \\\hline
\end{tabular}
\end{center}
\end{table}

\subsubsection{Experiment with More Languages (Chinese, Korean and Japanese)}
\label{sec:moreexp}

We employ wikipedia datasets of Chinese, Korean and Japanese together with the 11 languages mentioned before to  and also use the upsampling technique to make the dataset more balanced. Since there are no Korean and Japanese XNLI tasks, we only add Chinses (Zh) XNLI tasks in the downstream tasks evaluation. The result is as shown in Figure~\ref{tab:14lang}. From Table~\ref{tab:ta}, we observe that the improvements of NEZHA (BBPE) over others on De, Th and Zh are still notable. But we observe that NEZHA (BBPE) trained on 14 languages has notable performance drop on Ar, Ru compared with NEZHA (BBPE) trained on 11 languages as shown in Table~\ref{tab:bpe}. This is because the added three languages (Chinese, Korean and Japanese) have less sharing with Ar and Ru but consume several tokens in the vocabulary and this makes the tokens in Ar and Ru less. 

\begin{table}[ht]
\caption{Results on \emph{XNLI} (14 Pre-trained Languages)}
\label{tab:14lang}
\begin{center}
\begin{tabular}{|c|c|c|c|c|c|c|c|c|c|c|}
\hline
Model &  En     & Ar     & De     & Es     & Fr     & Ru     & Th & Zh  & Avg.  & Vocab\\ \hline
BERT  (google)          &       82.4    &72.0    &76.1    &78.4    &76.9    &74.5    &67.1 & 76.6  & 75.5  & ~120k  \\ \hline
 NEZHA  & 82.78    & 75.9    &79.1    &81.1    &79.3    &76.3    &68.8  & 76.8 & 77.5 & ~120k  \\ \hline
 NEZHA (BPE) & 80.7  & 70.4 & 77.5  & 78.3  & 78.6  & 72.9  & 67.5 & 77.2 & 75.4 & ~100k \\ \hline 
NEZHA (BBPE) & 82.4 & 72.1 & 79.6 & 80.2 & 79.3 & 75.1 & 70.2 & 77.8 & 77.1 & ~100k  \\ \hline
\end{tabular}
\end{center}
\end{table}

We also compare the BBPE vocabulary on 14 languages with Google multilingual vocabulary. Figure~\ref{fig:14lang} shows the language-wise comparison between Google multilingual vocabulary and our BBPE vocabulary. Each bar in the figure shows the relative difference of the tokens in the corresponding language (i.e., (A - B) / A, where A (B resp.) is the number of tokens in the language in BBPE (Google multilingual resp.) vocabulary). We observe that BBPE vocabulary has more tokens in Spanish, Arabic, Portuguese, Malay and Thai and less tokens in other languages compared with Google multilingual vocabulary. This implies that our BBPE vocabulary removes redundant tokens (esp., rare words/subwords) in languages like English, Russian, French etc. (esp. Japanese, Korean and Chinese which contains many rare characters/words) and reserves the tokens for low-resource languages like Arabic, Malay and Thai. 

\section{Conclusion}
\label{sec:conl}

In this technical report, we deployed the byte-level byte pair encoding (BBPE) in vocabulary building for the multilingual pre-trained language models. We detail several important techniques in developing byte-level vocabulary and provides insights on its practice. Our experiment shows that our method significantly outperforms several baselines which trains the models in text-level only (including Google's multilingual BERT, vanilla NEZHA and NEZHA with text-level BPE) and verified the effect of byte-level techniques that we deploy. 

\bibliographystyle{unsrt}  
\bibliography{references}  




\end{document}